\documentclass[runningheads]{llncs}
\usepackage{graphicx}

\usepackage{tikz}
\usepackage{comment} 
\usepackage{amsmath,amssymb} 
\usepackage{color}

\usepackage{comment}
\usepackage{amsmath,amssymb} 
\usepackage{color}
\usepackage{epsfig}
\usepackage{threeparttable}
\usepackage{multirow}
\usepackage{longtable}
\usepackage{rotating}
\usepackage{tabularx}
\usepackage{enumitem}
\usepackage{bbm}
\usepackage{pifont}
\usepackage{amssymb}
\usepackage{arydshln}
\usepackage{indentfirst}  

\usepackage{amsfonts}

\usepackage{booktabs}
\usepackage{mathrsfs}
\usepackage{wrapfig}

\newlength\savewidth\newcommand\shline{\noalign{\global\savewidth\arrayrulewidth
		\global\arrayrulewidth 1pt}\hline\noalign{\global\arrayrulewidth\savewidth}}

\definecolor{Olive_Green}{rgb}{0.0, 0.55, 0.0}
\usepackage[pagebackref=false,breaklinks=true,letterpaper=true,colorlinks,urlcolor = black,  citecolor = Olive_Green, bookmarks=false]{hyperref}

\usepackage{xspace}

\makeatletter
\DeclareRobustCommand\onedot{\futurelet\@let@token\@onedot}
\def\@onedot{\ifx\@let@token.\else.\null\fi\xspace}
\def\eg{\emph{e.g}\onedot} 
\def\ie{\emph{i.e}\onedot} 
 
\def\etc{\emph{etc}\onedot} \def\vs{\emph{vs}\onedot}

\def\@fnsymbol#1{\ensuremath{\ifcase#1\or *\or \dagger\or \ddagger\or
		\mathsection\or \mathparagraph\or \|\or **\or \dagger\dagger
		\or \ddagger\ddagger \else\@ctrerr\fi}}

\begin{document}
\pagestyle{headings}
\mainmatter
\def\ECCVSubNumber{3872}  

\title{Ocean: Object-aware Anchor-free Tracking} 

%

\titlerunning{Ocean: Object-aware Anchor-free Tracking}
%

\author{
	Zhipeng Zhang\textsuperscript{1 \thanks{\scriptsize Work performed when Zhipeng was an intern of Microsoft Research. $\dagger$ Corresponding author.}} 
	Houwen Peng\textsuperscript{2  $\dagger$}
	Jianlong Fu\textsuperscript{2}
	Bing Li\textsuperscript{1} 
	Weiming Hu\textsuperscript{1}  
	}

%
\authorrunning{ }
%

\institute{\;  $^{1}$CASIA \& AI School, UCAS \quad \quad \quad  \; $^{2} $Microsoft Research \quad \quad \quad \quad  \\
\email{zhipeng.zhang.cv@outlook.com \quad
	houwen.peng@microsoft.com}}

\maketitle

\begin{abstract}
Anchor-based Siamese trackers have achieved  remarkable advancements in accuracy, yet the further improvement is restricted by the lagged tracking robustness. We find the underlying reason is that the regression network in anchor-based methods is only trained on the positive anchor boxes (\ie, \emph{$IoU\geq0.6$}). This mechanism makes it difficult to refine the anchors whose overlap with the target objects are small. In this paper, we propose a novel object-aware anchor-free network to address this issue. First, instead of refining the reference anchor boxes, we directly predict the position and scale of target objects in an anchor-free fashion. Since each pixel in groundtruth boxes is well trained, the tracker is capable of rectifying inexact predictions of target objects during inference. Second, we introduce a feature alignment module to learn an object-aware feature from predicted bounding boxes. The object-aware feature can further contribute to the classification of target objects and background. Moreover, we present a novel tracking framework based on the anchor-free model. The experiments show that our anchor-free tracker achieves state-of-the-art performance on five benchmarks, including VOT-2018, VOT-2019, OTB-100, GOT-10k and LaSOT. The source code is available at \url{https://github.com/researchmm/TracKit}.

\end{abstract}

%
%
%
%

\section{Introduction}
Object tracking is a fundamental vision task. It aims to infer the location of an arbitrary target in a video sequence, given only its location in the first frame. The main challenge of tracking lies in that the target objects may undergo heavy occlusions, large deformation and illumination variations \cite{OTB-2015,SURVEY2}. Tracking at real-time speeds has a variety of applications, 
such as surveillance, robotics, autonomous driving and human-computer interaction \cite{intro1,intro2,intro3}.

In recent years, Siamese tracker has drawn great attention because of its balanced speed and accuracy. The seminal works, \ie, SINT~\cite{SINT} and SiamFC~\cite{siamFC}, employ Siamese networks to learn a similarity metric between the object target and candidate image patches, thus modeling the tracking as a search problem of the target over the entire image.
A large amount of follow-up Siamese trackers have been proposed and achieved promising performances \cite{CRPN,DSiam,SiamRPN++,siamRPN,SiamDW}. 
Among them, the Siamese region proposal networks, dubbed SiamRPN~\cite{siamRPN}, is representative. 
It introduces region proposal networks~\cite{FasterRCNN}, which consist of a classification network for foreground-background estimation and a regression network for anchor-box refinement, \ie, learning 2D offsets to the predefined anchor boxes. This anchor-based trackers have shown tremendous potential in tracking accuracy. However, since the regression network is only trained on the positive anchor boxes (\ie, $IoU \geq0.6$), it is difficult to refine the anchors whose overlap with the target objects are small. This will cause tracking failures especially when the classification results are not reliable.
For instance, due to the error accumulation in tracking, the predictions of target positions may become unreliable, \eg, $IoU<0.3$. The regression network is incapable of rectifying this weak prediction because it is previously unseen in the training set. 
As a consequence, the tracker gradually drifts in subsequent frames. 

\begin{figure}[t]
		\vspace{-1.3em}
	\centering
	\begin{minipage}[c]{.45\textwidth}
		\begin{flushleft}
			\caption{A comparison of the performance and speed of state-of-the-art tracking methods on VOT-2018. We visualize the Expected Average Overlap (EAO) with respect to the Frames-Per-Seconds (FPS). \emph{Offline-1} and \emph{Offline-2} indicate the proposed offline trackers with and without feature alignment module, respectively.}\label{VOT18FIG}
		\end{flushleft}
	\end{minipage}%
	\hspace{0.1em}
	\begin{minipage}[c]{.5\textwidth}
		\centering
		\begin{flushright}
			\includegraphics[width=1\textwidth,height=0.74\textwidth]{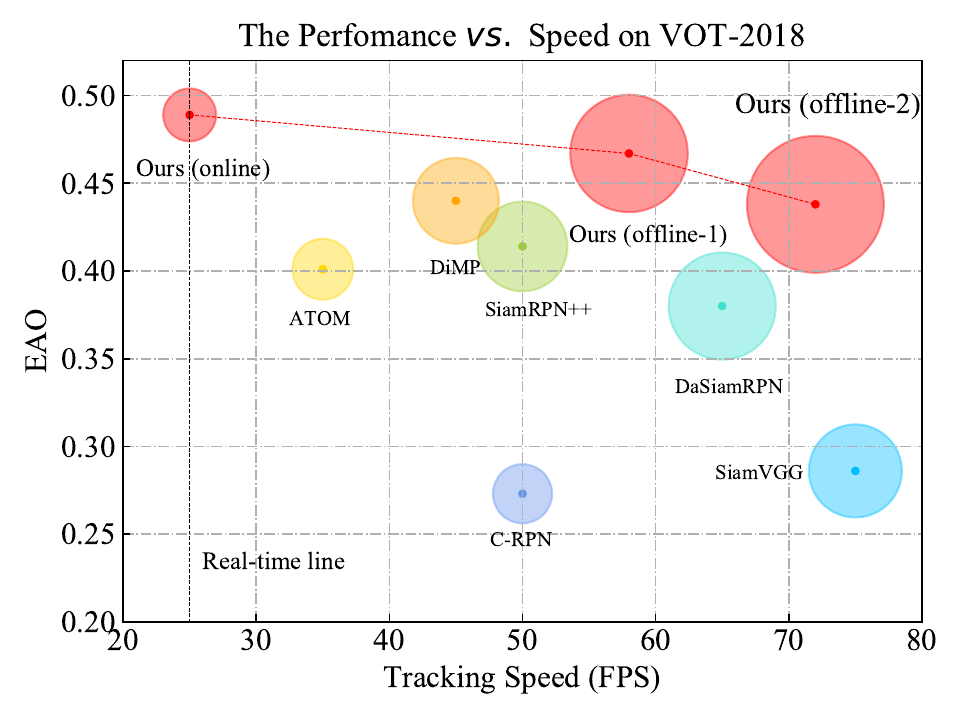}
		\end{flushright}
	\end{minipage}
		\vspace{-3em}
\end{figure}

It is natural to throw a question: \emph{can we design a bounding-box regressor with the capability of rectifying inaccurate predictions?} In this work, we show the answer is affirmative by proposing a novel object-aware anchor-free tracker. 
Instead of predicting the small offsets of anchor boxes, our object-aware anchor-free tracker directly regresses the positions of target objects in a video frame. More specifically, the proposed tracker consists of two components: an object-aware classification network and a bounding-box regression network. The classification is in charge of determining whether a region belongs to foreground or background, while the regression aims to predict the distances from each pixel within the target objects to the four sides of the groundtruth bounding boxes. Since each pixel in the groundtruth box is well trained, the regression network is able to localize the target object even when only a small region is identified as the foreground. Eventually, during inference, the tracker is capable of rectifying the weak predictions whose overlap with the target objects are small.

When the regression network predicts a more accurate bounding box (\eg, rectifying weak predictions), the corresponding features can in turn help the classification of foreground and background. We use the predicted bounding box as a reference to learn an object-aware feature for classification. 
More concretely, we introduce a feature alignment module, which contains a 2D spatial transformation to align the feature sampling locations with predicted bounding boxes (\ie, regions of candidate objects). This module guarantees the sampling is specified within the predicted regions, accommodating to the changes of object scale and position. 
Consequently, the learned features are more discriminative and reliable for classification.



The effectiveness of the proposed framework is verified on five benchmarks: VOT-2018 \cite{VOT-2018}, VOT-2019 \cite{VOT2019}, OTB-100 \cite{OTB-2015}, GOT-10k \cite{GOT10K} and LaSOT \cite{LASOT}. Our approach achieves state-of-the-art performance (an EAO of 0.467) on VOT-2018~\cite{VOT-2018}, while running at 58~\emph{fps}, as shown in Fig. \ref{VOT18FIG}. It obtains up to 92.2\% and 12.8\% relative improvements over the anchor-based methods, \ie, SiamRPN~\cite{siamRPN} and SiamRPN++~\cite{SiamRPN++}, respectively. On other datasets, the performance of our tracker is also competitive, compared with recent state-of-the-arts. In addition, we further equip our anchor-free tracker with a plug-in online update module, and enable it to capture the appearance changes of objects during inference. 
The online module further enhances the tracking performance, which shows the scalability of the proposed anchor-free tracking approach.

The main contributions of this work are two-fold.  1) We propose an object-aware anchor-free network based on the observation that the anchor-based method is difficult to refine the anchors whose overlap with the target object is small. The proposed algorithm can not only rectify the imprecise bounding-box predictons, but also learn an object-aware feature to enhance the matching accuracy. 2) We design a novel tracking framework by combining the proposed anchor-free network with an efficient feature combination module. 
The proposed tracking model achieves state-of-the-art performance on five benchmarks while running in real-time speeds.

\vspace{-0.8em}
\section{Related Work} \label{Sec2}
\vspace{-0.5em}
In this section, we review the related work on anchor-free mechanism 
and feature alignment in both tracking and detection, as well as briefly review recent Siamese trackers.


\textbf{Siamese trackers.} The pioneering works, \ie, SINT~\cite{SINT} and SiamFC~\cite{siamFC}, employ Siamese networks to offline train a similarity metric between the object target and candidate image patches. SiamRPN~\cite{siamRPN} improves it with a region proposal network, which amounts to a target-specific anchor-based detector. With the predefined anchor boxes, SiamRPN~\cite{siamRPN} can capture the scale changes of objects effectively. The follow-up studies mainly fall into two camps: designing more powerful backbone networks~\cite{SiamRPN++,SiamDW} or proposing more effective proposal networks~\cite{CRPN}. Although these offline Siamese trackers have achieved very promising results, their tracking robustness is still inferior to the recent state-of-the-art online trackers, such as ATOM~\cite{ATOM} and DiMP~\cite{DiMP}.

\textbf{Anchor-free mechanism.} Anchor-free approaches recently became popular in object detection tasks, because of their simplicity in architectures and superiority in performance~\cite{CenterNet,CornerNet,FCOS}. 
Different from anchor-based methods which estimate the offsets of anchor boxes, anchor-free mechanisms predict the location of objects in a direct way. 
The early anchor-free work~\cite{Unitbox} predicts the intersection over union with objects, while recent works focus on estimating the keypoints of objects, \eg, the object center~\cite{CenterNet} and corners~\cite{CornerNet}. Another branch of anchor-free detectors~\cite{YOLO,FCOS} predicts the object bounding box at each pixel, without using any references, \eg, anchors or keypoints. The anchor-free mechanism in our method is inspired by, but different from that in the recent detection algorithm~\cite{FCOS}. We will discuss the key differences in Sec.~\ref{Sec3.4}. 

\textbf{Feature alignment.} The alignment between visual features and reference ROIs (Regions of Interests) is vital for localization tasks, such as detection and tracking~\cite{CascadeRPN}. For example, 
ROIAlign~\cite{MASKRCNN} are commonly recruited in object detection to align the features with the reference anchor boxes, leading to remarkable improvements on localization precision. In visual tracking, there are also several approaches~\cite{RT-MDNet,SPM} considering the correspondence between visual features and candidate bounding boxes. However, these approaches only take account of the bounding boxes with high classification scores. If the high scores indicate the background regions, then the corresponding features will mislead the detection of target objects. To address this, we propose a novel feature alignment method, in which the alignment is independent of the classification results. We sample the visual features from the predicted bounding boxes directly, without considering the classification score, generating object-aware features. This object-aware features, in turn, help the classification of foreground and background.

\vspace{-0.5em}
\section{Object-aware Anchor-Free Networks} \label{Sec3}
\vspace{-0.3em}
This section proposes the \textbf{O}bje\textbf{c}t-awar\textbf{e} \textbf{a}nchor-free \textbf{n}etworks (Ocean) for visual tracking. The network architecture consists of two components: an object-aware classification network for foreground-background probability prediction and a regression network for target scale estimation. 
The input features to these two networks are generated by a shared backbone network (elaborated in Sec.~\ref{Sec4.1}). 
We introduce the regression network first, followed by the classification branch, because the regression branch provides object scale information to enhance the classification of the target object and background.


\vspace{-1em}
\subsection{Anchor-free Regression Network}  \label{Sec3.1}


Revisiting recent anchor-based trackers~\cite{SiamRPN++,siamRPN}, we observed that the trackers drift speedily when the predicted bounding box becomes unreliable. The underlying reason is that, during training, these approaches only consider the anchor boxes whose $IoU$ with groundtruth are larger than a high threshold, \ie, $\emph{IoU} \geq 0.6$. Hence, these approaches lack the competence to amend the weak predictions, \eg, the boxes whose overlap with the target are small. 


To remedy this issue, we introduce a novel anchor-free regression for visual tracking. It considers all the pixels in the groundtruth bounding box as the training samples. The core idea is to  estimate the distances from each pixel within the target object to the four sides of the groundtruth bounding box. Specifically, let $B = (x_{0}, y_{0}, x_{1}, y_{1}) \in \mathbb{R}^{4}$ denote the top-left and bottom-right corners of the groundtruth bounding box of a target object. A pixel is considered as the regression sample if its coordinates $(x, y)$ fall into the groundtruth box $B$. Hence, the labels $T^{*} = (l^{*}, t^{*}, r^{*}, b^{*})$ of training samples are calculated as

\begin{figure}[!t]
	
	\begin{center}
		
		
		\includegraphics[width=0.8\linewidth]{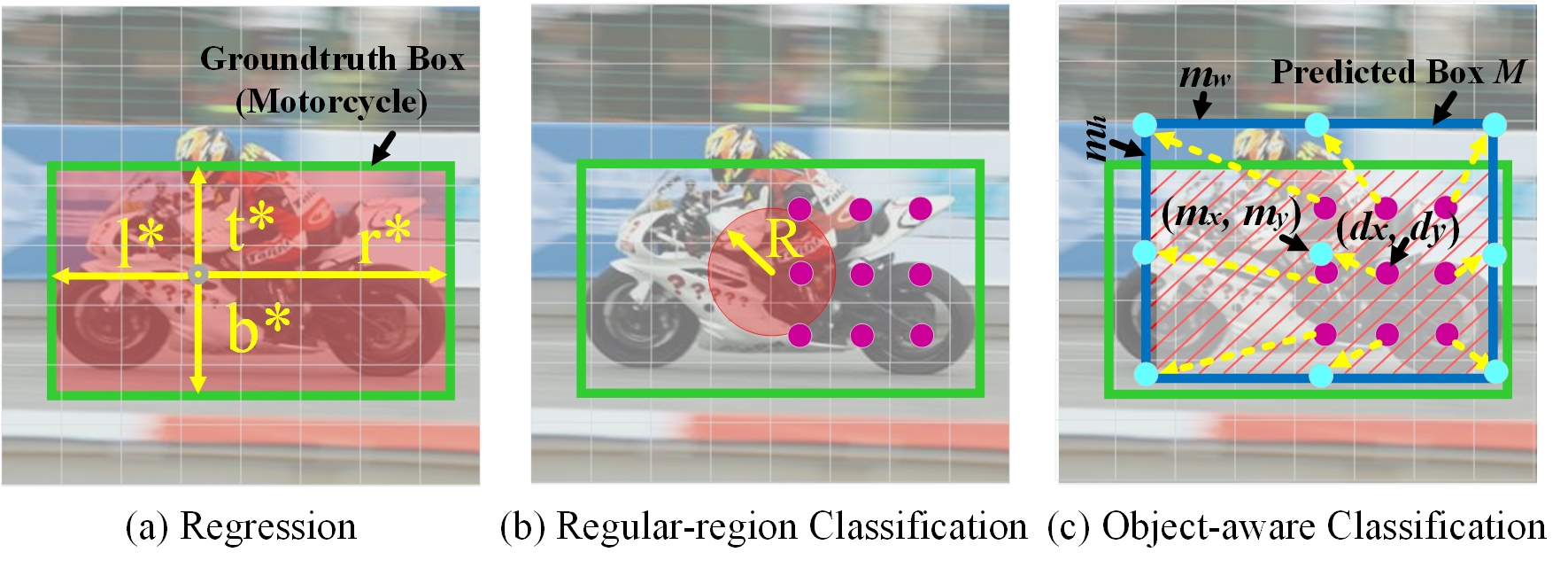}
		\vspace{-0.8em}
		
		\caption{(a) Regression: the pixels in groundtruth box, \ie the \textcolor{red}{red} region, are labeled as the positive samples in training. (b) Regular-region classification: the pixels closing to the target's center, \ie the \textcolor{red}{red} region, are labeled as the positive samples. The \textcolor[RGB]{204,0,153}{purple} points indicate the sampled positions of a location in the score map. (c) Object-aware classification: the \textit{IoU} of predicted box and groundtruth box, \ie, the region with \textcolor{red}{red} slash lines, is used as the label during training. The \textcolor[RGB]{102,255,255}{cyan} points represent the sampling positions for extracting object-aware features. The \textcolor{yellow}{yellow} arrows indicate the offsets induced by spatial transformation. Best viewed in color.} \label{SAMPLEFIG}
	\end{center}
	\vspace{-3.1em}
\end{figure}

\vspace{-0.8em}

\begin{equation} \label{Eq1}
\setlength{\abovedisplayskip}{4pt}
\setlength{\belowdisplayskip}{4pt}
\begin{split}
l^{*} = x - x_{0}&, t^{*} = y - y_{0},  \\
r^{*} = x_{1} - x&, b^{*} = y_{1} - y,  
\end{split}
\end{equation}
which represent the distances from the location $(x, y)$ to the four sides of the bounding box $B$, as shown in Fig.~\ref{SAMPLEFIG}(a). The learning of the regression network is through four $3\times3$ convolution layers with channel number of 256, followed by one $3\times 3$ layer with channel number of 4 for predicting the distances. As shown in Fig. \ref{FRAMEWORKFIG}, the upper ``Conv'' block indicates the regression network.

This anchor-free regression allows for all the pixels in the groundtruth box during training, thus it can predict the scale of target objects even when only a small region is identified as foreground. Consequently, the tracker is capable of rectifying weak predictions during inference to some extent. 

\vspace{-0.5em}
\subsection{Object-aware Classification Network} \label{Sec3.2}

In prior Siamese tracking approaches \cite{siamFC,SiamRPN++,siamRPN}, the classification confidence is estimated by the feature sampled from a fixed regular region in the feature map, \eg, the purple points in Fig. \ref{SAMPLEFIG}(b). This sampled feature depicts a fixed local region of the image, and it is not scalable to the change of object scale. As a result, the classification confidence is not reliable in distinguishing the target object from complex background.

To address this issue, we propose a feature alignment module to learn an object-aware feature for classification. The alignment module transforms the fixed sampling positions of a convolution kernel to align with the predicted bounding box. Specifically, for each location $(d_x, d_y)$ in the classification map, it has a corresponding object bounding box $M=(m_x, m_y, m_w, m_h)$ predicted by the regression network, where $m_x$ and $m_y$ denote the box center while $m_w$ and $m_h$ represent its width and height. Our goal is to estimate the classification confidence for each location $(d_x, d_y)$ by sampling features from the corresponding candidate region $M$. The standard 2D convolution with kernel size of $k \times k$ samples features using a fixed regular grid $\mathcal{G}= \{(-\left \lfloor k/2 \right \rfloor, -\left \lfloor k/2 \right \rfloor),..., (\left \lfloor k/2 \right \rfloor, \left \lfloor k/2 \right \rfloor)\}$, where $\left \lfloor \cdot \right \rfloor$ denotes the floor function. The regular grid $\mathcal{G}$ cannot guarantee the sampled features cover the whole content of region $M$. 

Therefore, we propose to equip the regular sampling grid $\mathcal{G}$ with a spatial transformation $\mathcal{T}$ to convert the sampling positions from the fixed region to the predicted region $M$. 
As shown in Fig. \ref{SAMPLEFIG}(c), the transformation $\mathcal{T}$ (the dashed yellow arrows) is obtained by measuring the relative direction and distance from the sampling positions in $\mathcal{G}$ (the purple points) to the positions aligned with the predicted bounding box (the cyan points). With the new sampling positions, the object-aware feature is extracted by the feature alignment module, which is formulated as

\begin{equation}
\setlength{\abovedisplayskip}{6pt}
\setlength{\belowdisplayskip}{6pt}
{\bf{f}}[u] = \underset{g\in \mathcal{G},\Delta t \in \mathcal{T}}{\sum } {\mathbf{w}}[g] \cdot {\bf{x}}[u+g+ \Delta t],
\label{Eq2}
\end{equation}
where $\mathbf{x}$ represents the input feature map, $\mathbf{w}$ denotes the learned convolution weight, $u$ indicates a location on the feature map, and ${\bf{f}}$ 
represents the output object-aware feature map. The spatial transformation $\Delta t \in \mathcal{T}$ represents the distance vector from the original regular sampling points to the new points aligned with the predicted bounding box. The transformation is defined as
\begin{equation}
\label{EqDELTA}
\mathcal{T} = \{(m_x, m_y) + \mathcal{B}\} - \{(d_x, d_y) + \mathcal{G}\},
\end{equation}
where $\{(m_x, m_y) + \mathcal{B}\}$ represents the   
sampling positions aligned with $M$, \eg, the cyan points in Fig. \ref{SAMPLEFIG}(c),  $\{(d_x, d_y) + \mathcal{G}\}$ indicates the 
regular sampling positions used in standard convolution, \eg, the purple points in Fig. \ref{SAMPLEFIG}(c), and $\mathcal{B} = \{(-m_w/2,-m_h/2),...,(m_w/2,m_h/2)\}$ denotes the coordinates of the new sampling positions (\eg, the cyan points in Fig. \ref{SAMPLEFIG}(c)) relative to the box center (\eg, $(m_x, m_y)$).
It is worth noting that when the transformation $\Delta t \in \mathcal{T}$ is set to $0$ in Eq.~(\ref{Eq2}), the feature sampling mechanism is degenerated to the fixed sampling on regular points, generating the regular-region feature.  
The transformations of the sampling positions are adaptive to the variations
of the predicted bounding boxes in video frames. Thus, the extracted object-aware
feature is robust to the changes of object scale, which is beneficial for feature matching during tracking. Moreover, the object-aware feature provides
a global description of the candidate targets, which enables the distinguish of the object and background to be more reliable.

We exploit both the object-aware feature and the regular-region feature to predict whether a 
region belongs to target object or image background. 
For the classification based upon the object-aware feature, we apply a standard convolution with kernel size of $3 \times3$ over ${\bf{f}}$ to predict the confidence $p_o$ (visualized as the ``OA.Conv" block of the classification network in Fig. \ref{FRAMEWORKFIG}). For the classification based on the regular-region feature, four $3\times 3$ standard convolution layers with channel number of 256, followed by one standard $3\times 3$ layer with channel number of one are performed over the regular-region feature ${\bf{f}}'$ to predict the confidence $p_r$ (visualized as the ``Conv" block of the classification network in Fig. \ref{FRAMEWORKFIG}). Calculating the summation of the confidence $p_o$ and $p_r$ obtains the final classification score. The object-aware feature provides a global description of the target, thus enhancing the matching accuracy of candiate regions. Meanwhile, the regular-region feature concentrates on local parts of images, which is robust to localize the center of target objects. The combination of the two features improves the reliability of the classification network.

\vspace{-0.5em}
\subsection{Loss Function}
\vspace{-0.3em}
To optimize the proposed anchor-free networks, we employ $IoU$ loss~\cite{Unitbox} and binary cross-entropy (BCE) loss~\cite{CrossEntropy} to train the regression and classification networks jointly. In regression, the loss is defined as
\begin{equation}
\setlength{\abovedisplayskip}{5pt}
\setlength{\belowdisplayskip}{5pt}
\mathcal{L}_{reg} = -\sum\nolimits_{i} ln (IoU(p_{reg}, T^{*})),
\label{Eq4}
\end{equation}
where $p_{reg}$ denotes the prediction, and $i$ indexes the training samples. In classification, the loss $\mathcal{L}_{o}$ based upon the object-aware feature ${\bf{f}}$ is formulated as
\begin{equation}
\setlength{\abovedisplayskip}{5pt}
\setlength{\belowdisplayskip}{5pt}
\mathcal{L}_{o} = -\sum\nolimits_{j} p_{o}^{*}log(p_{o}) + (1-p_{o}^{*})log(1-p_{o}),
\label{Eq5}
\end{equation}
while the loss $\mathcal{L}_{r}$ based upon the regular-region feature ${\bf{f}}'$ is formulated as
\begin{equation}
\setlength{\abovedisplayskip}{5pt}
\setlength{\belowdisplayskip}{5pt}
\mathcal{L}_{r} = -\sum\nolimits_{j} p_{r}^{*}log(p_{r}) + (1-p_{r}^{*})log(1-p_{r}),
\label{Eq6}
\end{equation}
where $p_{o}$ and $p_{r}$ are the classification score maps computed over the object-aware feature and regular-region feature respectively, $j$ indexes the training samples for classification, and $p_o^{*}$ and $p_r^{*}$ denote the groundtruth labels. More concretely, $p_o^{*}$ is a probabilistic label, in which each value indicates the \textit{IoU} between the predicted bounding box and groundtruth, \ie, the region with red slash lines in Fig. \ref{SAMPLEFIG}(c).  $p_r^{*}$ is a binary label, where the pixels closing to the center of the target are labeled as 1, \ie, the red region in Fig. \ref{SAMPLEFIG}(b), which is formulated as

\begin{equation} \label{Eq3}
\setlength{\abovedisplayskip}{0pt}
\setlength{\belowdisplayskip}{4pt}
p_{r}^{*}[v]=\left\{
\begin{array}{lr}
1, \: \: if ||v-c|| \leq R,   \\
0, \: \: otherwise.
\end{array}
\right.
\end{equation}

The joint training of the entire object-aware anchor-free networks is to optimize the following objective function:
\begin{equation}
\setlength{\abovedisplayskip}{4pt}
\setlength{\belowdisplayskip}{4pt}
\mathcal{L} = \mathcal{L}_{reg} + \lambda_{1} \mathcal{L}_{o} + \lambda_{2} \mathcal{L}_{r},
\label{Eq7}
\end{equation}
where $\lambda_{1}$ and $\lambda_{2}$ are the tradeoff hyperparameters.

\vspace{-0.7em}
\subsection{Relation to Prior Anchor-Free Work} \label{Sec3.4}
\vspace{-0.3em}
 
Our anchor-free mechanism shares similar spirit with recent detection methods~\cite{CenterNet,CornerNet,FCOS} (discussed in Sec.~\ref{Sec2}). 
In this section, we further discuss the differences to the most related work, \ie, FCOS~\cite{FCOS}.
Both FCOS and our method predict the object locations directly on the image plane at pixel level. 
However, our work differs from FCOS~\cite{FCOS} in two fundamental ways.
1) In FCOS~\cite{FCOS}, the training samples for the classification and regression networks are identical. Both are sampled from the positions within the groundtruth boxes. Differently, in our method, the data sampling strategies for classification and regression are asymmetric 
which is tailored for tracking tasks. More specifically, the classification network only considers the pixels closing to the target as positive samples (\ie, $R\leq16$ pixels), while the regression network considers all the pixels in the ground-truth box as training samples. This fine-grained sampling strategy guarantees the classification network can learn a robust similarity metric for region matching, which is important for tracking. 2) In FCOS \cite{FCOS}, the objectness score is calculated with the feature extracted from a fixed regular-region, similar to the purple points in Fig. \ref{SAMPLEFIG}(b). By contrast, our method additionally introduce an object-aware feature, which captures the global appearance of target objects. The object-aware feature aligns the sampling regions with the predicted bounding box (\eg, cyan points in Fig. \ref{SAMPLEFIG}(c)), thus it is adaptive to the scale change of objects. The combination of the regular-region feature and the object-aware feature allows the classification to be more reliable, as verified in Sec.~\ref{Sec5.3}.

\vspace{-1.3em}
\section{Object-aware Anchor-Free Tracking} \label{Sec4}
\vspace{-0.7em}
This section depicts the tracking algorithm building upon the proposed object-aware anchor-free networks (Ocean). It contains two  parts: an offline anchor-free model and an online update model, as illustrated in Fig. \ref{FRAMEWORKFIG}.

\vspace{-1.4em}
\subsection{Framework}
\vspace{-0.5em}
\label{Sec4.1}
The offline tracking is built on the object-aware anchor-free networks, consisting of three steps: feature extraction, combination and target localization.

\textbf{Feature extraction.} Following the architecture of Siamese tracker~\cite{siamFC}, our approach takes an image pair as input, \ie., an exemplar image and a candidate search image. The exemplar image represents the object of interest, \ie, an image patch centered on the target object in the first frame, while the search image is typically larger and represents the search area in subsequent video frames. Both inputs are processed by a modified ResNet-50~\cite{ResNet} backbone and then yield two feature maps. More specifically, we cut off the last stage of the standard ResNet-$50$~\cite{ResNet}, and only retain the first fourth stages as the backbone. The first three stages share the same structure as the original ResNet-$50$. In the fourth stage, the convolution stride of down-sampling unit~\cite{ResNet} is modified from 2 to 1 to increase the spatial size of feature maps, meanwhile, all the $3\times3$ convolutions are augmented with a dilation with stride of 2 to increase the receptive fields.
These modifications increase the resolution of output features, thus improving the feature capability on object localization~\cite{HoleConv,SiamRPN++}.


\begin{figure*}[!t]
	\begin{center}
		\vspace{-1.5em}
		
		
		\includegraphics[height=0.45\linewidth]{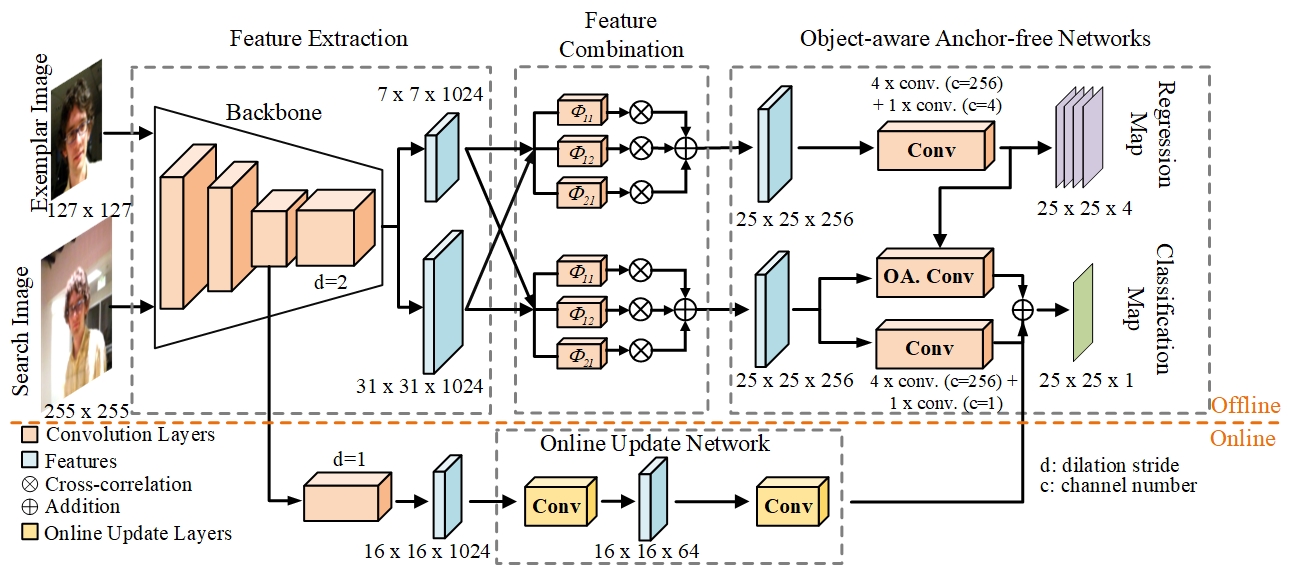}
		\vspace{-2.8em}
		
		\caption{
			Overview of the proposed tracking framework, consisting of an offline anchor-free part (top) and an online model update part (bottom). The offline tracking includes feature extraction, feature combination and target localization with object-aware anchor-free networks, as elaborated in Sec. \ref{Sec4.1}. The plug-in online update network models the appearance changes of target objects, as detailed in Sec. \ref{Sec4.2}.
			$\Phi_{ab}$ indicates a $3\times 3$ convolution layer with dilation stride of $a$ along the $X$-axis and $b$ along the $Y$-axis.
		}
		
		\label{FRAMEWORKFIG}
	\end{center}
	
	\vspace{-3.1em}
\end{figure*}

\textbf{Feature combination.}
This step exploits a depth-wise cross-correlation operation~\cite{SiamRPN++} to combine the extracted features of the exemplar and search images, and generates the corresponding similarity features for the subsequent target localization. Different from the previous works performing the cross-correlation on multi-scale features~\cite{SiamRPN++}, our method only performs over a single scale, \ie, the last stage of the backbone. We pass the single-scale features through three parallel dilated convolution layers~\cite{Dilated}, and then fuse the correlation features through point-wise summation, as presented in Fig.~\ref{FRAMEWORKFIG} (feature combination). 

For concreteness, the feature combination process can be formulated as 
\begin{equation} \label{Eq8}
\setlength{\abovedisplayskip}{5pt}
\setlength{\belowdisplayskip}{5pt}
S = \sum\nolimits_{ab}{\Phi_{ab}({\bf{f}}_e) * \Phi_{ab}({\bf{f}}_s)}
\end{equation}
where ${\bf{f}}_e$ and ${\bf{f}}_s$ represent the features of the exemplar and search images respectively, $\Phi{_{ab}}$ indicates a single dilated convolution layer, and $*$ denotes the cross-correlation operation~\cite{siamFC}. The  kernel size of the dilated convolution $\Phi{_{ab}}$ is set to $3\times3$, while the dilation strides are set to $a$ along the $X$-axis and $b$ along the $Y$-axis. $\Phi{_{ab}}$ also reduces the feature channels from $1024$ to $256$ to save computation cost. In experiments, we found that increasing the diversity of dilations can improve the representability of features, thereby we empirically choose three different dilations, whose strides are set to $(a,b) \in \{(1,1),(1,2),(2,1)\}$. The convolutions with different dilations can capture the features of regions with different scales, improving the scale invariance of the final combined features.

\textbf{Target localization.} This step employs the proposed object-aware anchor-free networks to localize the target from search images. The probabilities $p_o$ and $p_r$ predicted by the classification network are averaged with a weight $\omega$ as
\begin{equation}
\setlength{\abovedisplayskip}{6pt}
\setlength{\belowdisplayskip}{6pt}
p_{cls} = \omega p_{o} + (1 - \omega) p_{r}.
\label{Eq9}
\end{equation}
Similar to prior works~\cite{siamFC,SiamRPN++}, we impose a penalty on scale change to suppress the large variation of object size and aspect ratio as follows
\begin{equation}
\setlength{\abovedisplayskip}{6pt}
\setlength{\belowdisplayskip}{6pt}
\alpha = exp({k\cdot max(\frac{r}{{r}'},\frac{{r}'}{r})\cdot max(\frac{s}{{s}'},\frac{{s}'}{s})}),
\label{Eq10}
\end{equation}
where $k$ is a hyper-parameter, $r$ and $r'$ represent the aspect ratio of the predicted bounding boxes in the previous and current frames respectively, while $s$ and $s'$ denote the size (\ie, height and width) of the predicted boxes in the previous and current frames. The final target classification probability $\hat{p}_{cls}$ is calculated as $\hat{p}_{cls} = \alpha \cdot p_{cls}$. The maximum value in the classification map $\hat{p}_{cls}$ indicates the position of the foreground target. To keep the shape of predicted bounding boxes changing smoothly, a linear weight function is used to calculate the final scale as $\hat{s}_{reg}$ = $\beta\cdot{s}' + (1 - \beta)\cdot s$, where $\beta$ is a weight parameter.

\vspace{-0.5em}
\subsection{Integrating Online Update}
\label{Sec4.2}

We further equip the offline algorithm with an online update model. Inspired by~\cite{DiMP,ATOM}, we introduce an online branch to capture the appearance changes of target object during tracking. As shown in Fig. \ref{FRAMEWORKFIG} (bottom part), the online branch inherits the structure and parameters from the first three stages of the backbone network, \ie, modified ResNet-50~\cite{ResNet}. The fourth stage keep the same structure as the backbone, but its initial parameters are obtained through the pretraining strategy proposed in \cite{DiMP}. For model update, we employ the fast conjugate gadient algorithm~\cite{DiMP} to train online branch during inference. The foreground score maps estimated by the online branch and the classification branch are weighted as
\begin{equation}
\setlength{\abovedisplayskip}{4pt}
\setlength{\belowdisplayskip}{10pt}
p = \omega' p_{onl} + (1 - \omega') \hat{p}_{cls},
\label{Eq11}
\end{equation}
where $\omega'$ represents the weights between the classification score $\hat{p}_{cls}$ and the online estimation score $p_{onl}$. Note that the IoUNet in \cite{DiMP,ATOM} is not used in our model. We refer readers to \cite{DiMP,ATOM} for more details.


\vspace{-0.5em}
\section{Experiments}
\label{Sec5}
\vspace{-0.6em}
This section presents the results of our Ocean tracker on five tracking benchmark datasets, with comparisons to the state-of-the-art algorithms. 
Experimental analysis is provided to evaluate the effects of each component in our model.

\vspace{-0.5em}
\subsection{Implementation Details}

\textbf{Training.} The backbone network is initialized with the parameters pretrained on ImageNet \cite{ImageNet}. The proposed trackers are trained on the datasets of Youtube-BB \cite{YTB}, ImageNet VID \cite{ImageNet}, ImageNet DET \cite{ImageNet}, GOT-10k \cite {GOT10K} and COCO \cite{COCO}. The size of input exemplar image is $127\times 127$ pixels, while the search image is $255\times 255$ pixels. We use synchronized SGD \cite{SGD} on 8 GPUs, with each GPU hosting 32 images, hence the mini-batch size is 256 images per iteration. There are 50 epochs in total. Each epoch uses $6 \times 10 ^{5}$ training pairs. For the first 5 epochs, we start with a warmup learning rate of $10^{-3}$ to train the object-aware anchor-free networks, while freezing the parameters of the backbone. For the remaining epochs, the backbone network is unfrozen, and the whole network is trained end-to-end with a learning rate exponentially decayed from $5\times 10^{-3}$ to $10^{-5}$. The weight decay and momentum are set to $10^{-3}$ and 0.9, respectively. 
The threshold $R$ of the classification label in Eq. (\ref{Eq3}) is set to 16 pixels. The weight parameters $\lambda_{1}$ and $\lambda_{2}$ in Eq. (\ref{Eq7}) are set to 1 and 1.2, respectively. 

We noticed that the training settings (data selection, iterations, \etc) are often different in recent trackers, \eg, SiamRPN \cite{siamRPN}, SiamRPN++\cite {SiamRPN++}, ATOM \cite{ATOM} and DiMP \cite{ATOM}. It is difficult to compare different models under a unified training schedule. But for a fair comparison, we additionally evaluate our method and SiamRPN++~\cite{SiamRPN++} under the same training setting, as discussed in Sec. \ref{Sec5.3}.


\textbf{Testing.} For the offline model, tracking follows the same protocols as in \cite{siamFC,siamRPN}. The feature of the target object is computed once at the first frame, and then is continuously matched to subsequent search images. The fusion weight $\omega$ of the object-aware classification score in Eq. (\ref{Eq9}) is set to 0.07, while the weight $\omega'$ in Eq. (\ref{Eq11}) is set to 0.5. The hyperparameter $k$ in Eq. (\ref{Eq10}) for the penalty of large scale change is set to 0.021, while the scale weight $\beta$ is set to 0.7. These hyper-parameters in testing are selected with the tracking toolkit~\cite{SiamDW}, which contains an automated parameter tuning algorithm. 
Our trackers are implemented using Python 3.6 and PyTorch 1.1.0. The experiments are conducted on a server with 8 Tesla V100 GPUs and a Xeon E5-2690 2.60GHz CPU. Note that we run the proposed tracker three times, the standard deviation of the performance is $\pm 0.5\%$, demonstrating the stability of our model. We report the average performance of the three-time runs in the following comparisons.

\setlength{\tabcolsep}{.2em}

\begin{table*}[!t] \scriptsize
		\vspace{-0.6em}
	\centering
	\begin{tabular}{l| c c c c c c c c c c c}
		& CRPN &ECO & STRCF& LADCF & ATOM &SRCNN& SiamRPN++  & DiMP & \textbf{Ocean} & \textbf{Ocean}\\ %
		& \cite{CRPN} & \cite{ECO}  & \cite{STRCF}  & \cite{LADCF} & \cite{ATOM} & \cite{SiamRCNN} &\cite{SiamRPN++} & \cite{DiMP} & offline & online\\
		
		\shline %
		EAO $\uparrow$  & 0.273 &0.280  & 0.345 & 0.389 & 0.401 & 0.408 & 0.414 & \color{blue}\textbf{0.440} & \color{green}\textbf{0.467} & \color{red}\textbf{0.489} \\ %
		A $\uparrow$  & 0.550 & 0.484& 0.523 & 0.503 & 0.590 & \color{red}\textbf{0.609} & \color{green}\textbf{0.600} & 0.597  & \color{blue}\textbf{0.598} & 0.592\\ %
		R $\downarrow$  & 0.320 &0.276 &0.215 & \color{green}\textbf{0.159} & 0.204& 0.220 & 0.234 & \color{blue}\textbf{0.153} & 0.169 & \color{red}\textbf{0.117}\\ %
		
	\end{tabular}
	
	
	
	\caption{Performance comparisons on VOT-2018 benchmark. {\color{red}{Red}}, {\color{green}green} and {\color{blue}{blue}} fonts indicate the top-3 trackers. ``Ocean'' denotes our propose model.} 
	\label{VOT18TAB}
	\vspace{-2em}
\end{table*}

\setlength{\tabcolsep}{.2em}
\begin{table}[!t] \scriptsize
	\centering
	\begin{tabular}{c| c  c c c c c c cc c c}
		& MemDTC  & SiamMASK  & SiamRPN++ & ATOM  & STN & DiMP$^r$ & DiMP$^b$& \textbf{Ocean} & \textbf{Ocean}\\ %
		&\cite{MemDTC}  & \cite{SiamMASK} & \cite{SiamRPN++} & \cite{ATOM} & \cite{STN} & \cite{DiMP} & \cite{DiMP} & offline & online\\
		
		\shline %
		EAO $\uparrow$  &0.228   & 0.287 & 0.292 & 0.301 & 0.314 &0.321 &\color{red}\textbf{0.379} & \color{blue}\textbf{0.327} & \color{green}\textbf{0.350} \\ %
		A $\uparrow$ &0.485    & \color{green}\textbf{0.594} & 0.580 & \color{red}\textbf{0.603} & 0.589 & 0.582& \color{green}\textbf{0.594}& 0.590 & \color{green}\textbf{0.594}\\ %
		
		R $\downarrow$  &0.587 & 0.461 & 0.446 & 0.411  & \color{blue}\textbf{0.349}  & 0.371& \color{red}\textbf{0.278}&0.376 & \color{green}\textbf{0.316}\\ %

	\end{tabular}
	
	\caption{Performance comparisons on VOT-2019. The ``DiMP$^{r}$'' and ``DiMP$^{b}$'' indicate realtime and baseline performances of DiMP, as reported in \cite{VOT2019}.}
	\vspace{-2em}
	\label{VOT19TAB}
\end{table}

\setlength{\tabcolsep}{.25em}
\begin{table}[!t] \scriptsize
	\centering
	\begin{tabular}{l| c  c c cc c c c c c}
		& CFNet & MDNet &SiamFC &ECO& DSiam& SiamRPN++ & ATOM & DiMP &  \textbf{Ocean} & \textbf{Ocean}\\ %
		& \cite{CFNet} & \cite{MDNet} & \cite{siamFC}& \cite{ECO} & \cite{DSiam}& \cite{SiamRPN++} & \cite{ATOM} & \cite{DiMP} & offline & online \\
		
		\shline %
		AO $\uparrow$    & 0.261 & 0.299 & 0.392 & 0.395&  0.417 &0.518 & 0.556 & \color{red}\textbf{0.611} & \color{green}\textbf{0.592} & \color{red}\textbf{0.611} \\ %
		SR$_{0.5}$ $\uparrow$    &0.243 & 0.303 & 0.406&0.407&0.461 & 0.618 & 0.634 & \color{green}\textbf{0.712}  & \color{blue}\textbf{0.695} & \color{red}\textbf{0.721}\\ %

	\end{tabular}
	
	\caption{Performance comparisons on GOT-10k test set.}
	\vspace{-3.5em}
	\label{GOT10KTAB}
\end{table}

\textbf{Evaluation datasets and metrics.}\label{Sec5.1}
We use five benchmark datasets including VOT-2018 \cite{VOT-2018}, VOT-2019 \cite{VOT2019}, OTB-100 \cite{OTB-2015}, GOT-10k \cite{GOT10K} and LaSOT \cite{LASOT} for tracking performance evaluation. In particular, VOT-2018 \cite{VOT-2018} contains 60 sequences. VOT-2019 \cite{VOT2019} is developed by replacing the $20\%$ least challenging videos in VOT-2018~\cite{VOT-2018}. We adopt the Expected Average Overlap (EAO)~\cite{VOT2019} which takes both accuracy (A) and robustness (R) into account to evaluate overall performance. The standardized OTB-100 \cite{OTB-2015} benchmark consists of 100 videos. Two metrics, \ie, precision (Prec.) and area under curve (AUC) are used to rank the trackers. GOT-10k \cite{GOT10K} is a large-scale dataset containing over 10 thousand videos. The trackers are evaluated using an online server on a test set of 180 videos. It employs the widely used average overlap (AO) and success rate (SR) as performance indicators. Compared to these benchmark datasets, LaSOT \cite{LASOT} has longer sequences, with an average of 2,500 frames per sequence. Success (SUC) and precision (Prec.) are used to evaluate tracking performance.

\vspace{-1em}
\subsection{State-of-the-art Comparison}
\begin{wrapfigure}{r}{0.6\textwidth}
	
	\centering
	\begin{minipage}[c]{8cm}
		\hspace{-1.5em}
		\includegraphics[width=8cm]{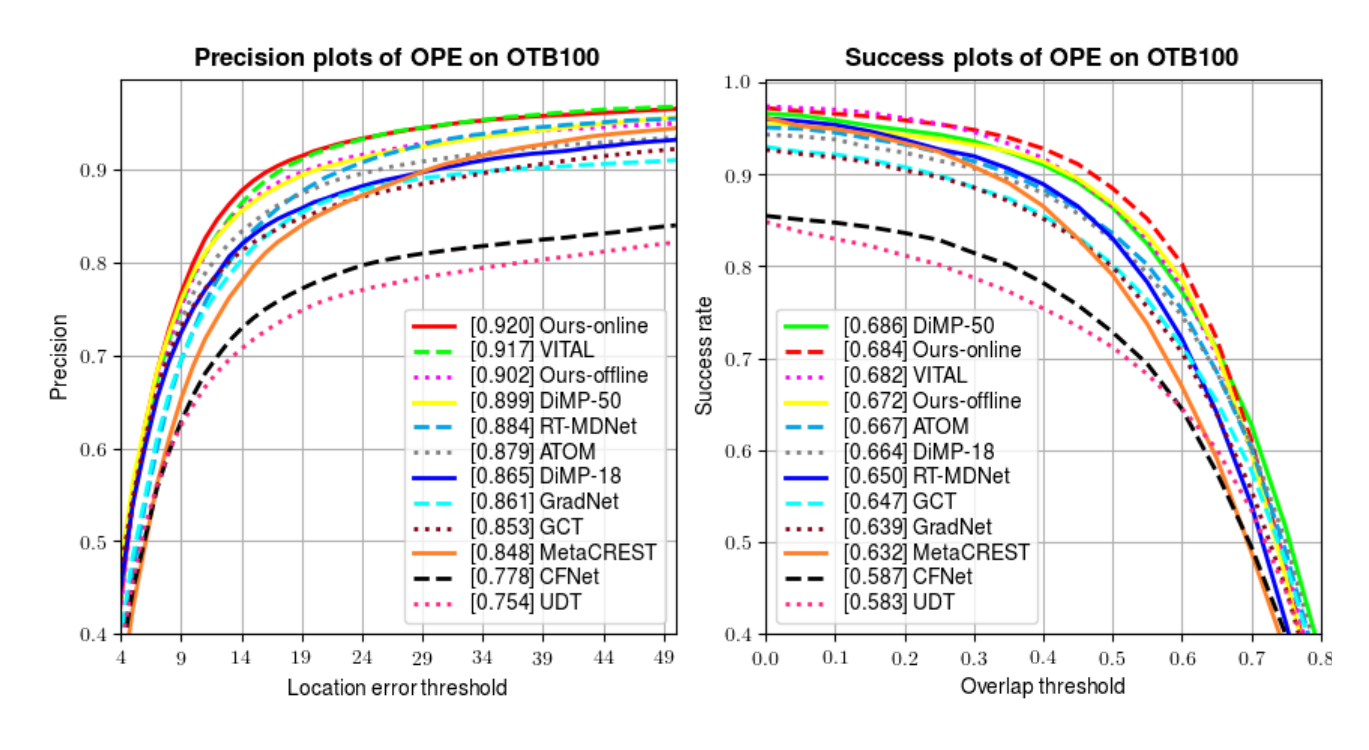}
	\end{minipage}%
	\hfill
	\vspace{-1em}
	\begin{minipage}[c]{8cm}
		\hspace{-1.5em}
		\includegraphics[width=8cm]{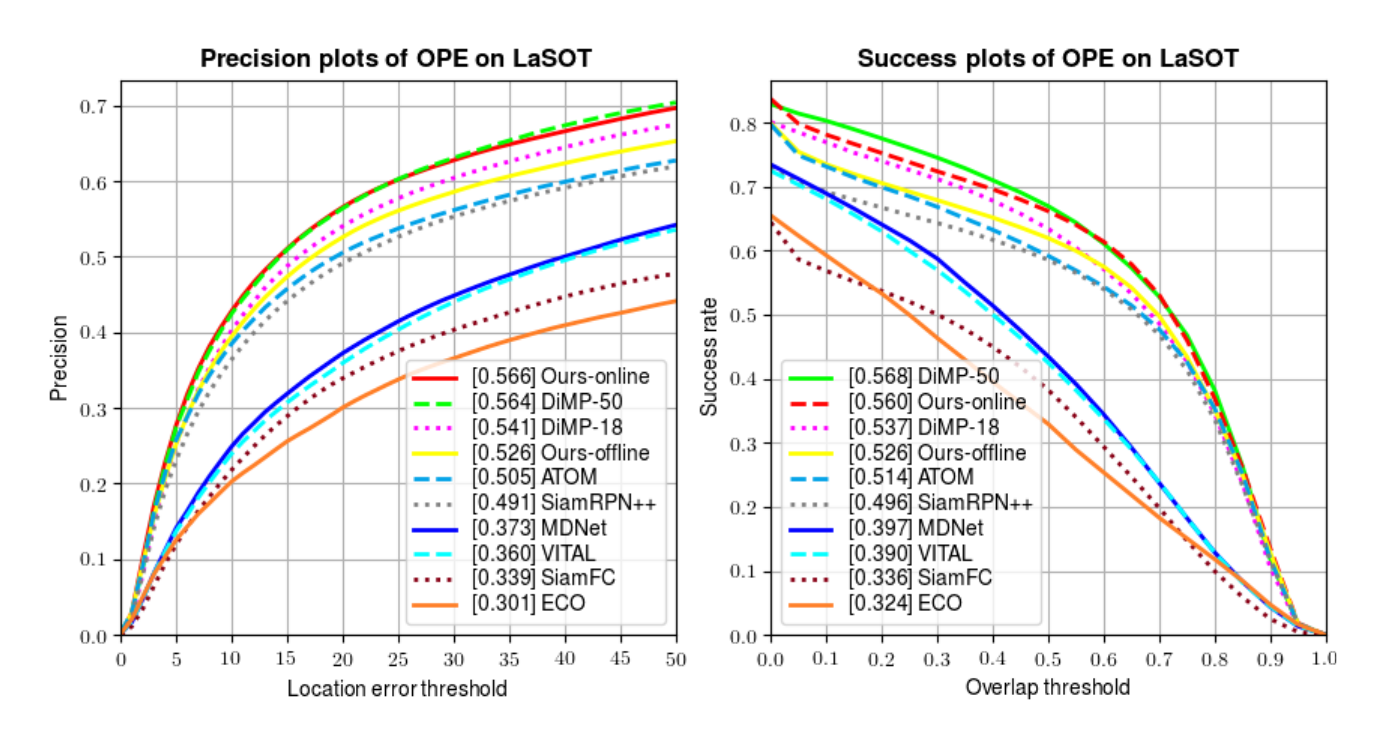}
	\end{minipage}%
	\hfill
	\begin{minipage}[c]{7cm}
		\vspace{-1em}
		\caption{Success and precision plots on OTB-100 \cite{OTB-2015} (top) and LaSOT \cite{LASOT} (bottom).}
		\label{OTBFIG}
	\end{minipage}%
	
	\vspace{-2em}
\end{wrapfigure}

To extensively evaluate the proposed method, we compare it with 22 state-of-the-art trackers, which cover most of current representative methods. There are 9 anchor-based Siamese framework based methods (SiamFC \cite{siamFC}, GradNet \cite{GradNet}, DSiam \cite{DSiam}, MemDTC \cite{MemDTC}, SiamRPN \cite{siamRPN}, C-RPN \cite{CRPN}, SiamMASK \cite{SiamMASK}, SiamRPN++ \cite{SiamRPN++} and SiamRCNN \cite{SiamRCNN}), 8 discriminative correlation filter based methods (CFNet \cite{CFNet}, ECO \cite{ECO}, STRCF \cite{STRCF}, LADCF \cite{LADCF}, UDT \cite{UDT}, STN \cite{STN}, ATOM \cite{ATOM} and DiMP \cite{DiMP}), 
3 multi-domain learning based methods (MDNet \cite{MDNet}, RT-MDNet \cite{RT-MDNet} and VITAL \cite{VITAL}), 1 graph network based method (GCT \cite{GCT}) and 1 meta-learning based tracker (MetaCREST \cite{METATRACKER}). 
The results are summarized in Tab. \ref{VOT18TAB} - \ref{GOT10KTAB} and Fig. \ref{OTBFIG}. 


\noindent\textbf{VOT-2018.} The evaluation on VOT-2018 is performed by the official toolkit \cite{VOT-2018}. As shown in Tab. \ref{VOT18TAB}, 
Our offline Ocean tracker outperforms the champion method of VOT-2018, \ie, (LADCF \cite{LADCF}), by 7.8 points. Compared to the state-of-the-art offline tracker SiamRPN++ \cite{SiamRPN++}, our offline model achieves EAO improvements of 5.3 points, while running faster, as shown in Fig.~\ref{VOT18FIG}. 
It is worth noting that the improvements mainly come from the robustness score, which obtains $27.8\%$ relative increases over SiamRPN++. Moreover, our offline model is  superior to the recent online trackers ATOM \cite{ATOM} and DiMP \cite{DiMP}. The online augmented model further improves our tracker by 2.2 points in terms of EAO.

\noindent\textbf{VOT-2019.} Tab. \ref{VOT19TAB} reports the evaluation results with the comparisons to recent prevailing trackers on VOT-2019. 
We can see that the recent proposed DiMP \cite{DiMP} achieves the best performance, while our method ranks second. However, in real-time testing scenarios, our offline Ocean tracker achieves the best performance, surpassing DiMP$^r$ by 0.6 points in terms of EAO. Moreover, the EAO of our offline model surpasses SiamRPN++ \cite{SiamRPN++} by 3.5 points.

\noindent\textbf{GOT-10k.} The evaluation on GOT-10k follows the protocols in \cite{GOT10K}. The proposed offline Ocean tracker model achieves the state-of-the-art AO score of 0.592, outperforming SiamRPN++ \cite{SiamRPN++}, as shown in Tab. \ref{GOT10KTAB}. Our online model improves the AO by 4.5 points over ATOM \cite{ATOM}, while outperforming DiMP~\cite{DiMP} by 0.9 points in terms of success rate.

\noindent\textbf{OTB-100.} The last evaluation in short-term tracking is performed on the classical OTB-100 benchmark. As reported in Fig.~\ref{OTBFIG}, among the compared methods, our online tracker achieves the best precision score of 0.920, while DiMP \cite{SiamRPN++} achieves best AUC score of 0.686.

\noindent\textbf{LaSOT.}  To further evaluate the proposed models, we report the results on LaSOT, which is larger and more challenging than previous benchmarks. The results on the 280 videos test set is presented in Fig. \ref{OTBFIG}. Our offline Ocean tracker achieves SUC score of 0.527, outperforming SiamRPN++ with score of 0.496. Compared to ATOM \cite{ATOM},
our online tracker improves the SUC score by 4.6 points, giving comparable results to top-ranked tracker DiMP-50 \cite{DiMP}. Moreover, the proposed online tracker achieves the best precision score of 0.566.

\begin{table}[!t]\selectfont
	\vspace{-0em}
	\begin{minipage}[t]{.5\textwidth}
		\setlength{\abovecaptionskip}{0pt}%
		\centering
		
		\fontsize{9.5pt}{4mm}
		\begin{threeparttable}
			\begin{tabular}{  @{}c@{}| @{}l@{} |@{}c@{}}
				
				\#Num & ~Components & ~EAO~
				\\  \cline{1-3}
				\textcircled{1} & ~baseline   & 0.358
				
				\\
				\textcircled{2} & ~$+$ centralized sampling~ & 0.396
				
				\\
				\textcircled{3} & ~$+$ feature combination ~  & 0.438
				
				\\
				\textcircled{4} & ~$+$ object-aware classification  & 0.467
				
				\\
				\textcircled{5} & ~$+$ online update ~ & \small{0.489}

			\end{tabular}
		\end{threeparttable}
		\caption{Component-wise analysis of our model. The results prove that each component is important in our framework.}
		\label{COMTAB}
		
	\end{minipage}\hfill
	\begin{minipage}[t]{.48\textwidth}%
		\setlength{\abovecaptionskip}{0pt}%
		\centering
		
		\fontsize{9.5pt}{4mm}
		\begin{threeparttable} 
			\begin{tabular}{  @{}c@{}| @{}l@{} |@{}c@{}}
				
				\#Num ~& ~Dilated Kernels~~~~~ & ~~EAO~~
				\\  \cline{1-3}
				\textcircled{1} & ~~$\Phi_{11}$  & 0.425  
				
				\\
				\textcircled{2} & ~ $\Phi_{11}\Phi_{11}$ & 0.433   
				
				\\
				\textcircled{3} & ~ $\Phi_{11}\Phi_{12}$ ~  & 0.446   
				
				\\
				\textcircled{4} & ~ $\Phi_{11}\Phi_{21}$  & 0.443  
				
				\\
				\textcircled{5} & ~ $\Phi_{11}\Phi_{12}\Phi_{21}$ & 0.467  

			\end{tabular}
		\end{threeparttable}
		\caption{Analysis of the impact of different strides over dilated convolution in the feature combination module.}
		
		\label{FCTAB}
	\end{minipage}
	\vspace{-3em}
	
\end{table}

\vspace{-0.8em}
\subsection{Analysis of the Proposed Method} \label{Sec5.3}


\noindent\textbf{Component-wise analysis.} To verify the efficacy of the proposed method, we perform a component-wise analysis on the VOT-2018 benchmark, as presented in Tab. \ref{COMTAB}. The baseline model consists of a backbone network (detailed in Sec. \ref{Sec4.1}), an anchor-free regression network (detailed in Sec. \ref{Sec3.1}) and a classification network using regular-region feature (detailed in Sec. \ref{Sec3.2}). In the training of baseline model, all pixels in the groundtruth box are considered as positive samples. The baseline model obtains an EAO of 0.358. In \textcircled{2}, the ``centralized sampling'' indicates that we only consider the pixels closing to the target's center as positive samples in the training of classification (formulated as Eq. (\ref{Eq3})). It brings significant gains, \ie, 3.8 points on EAO (\textcircled{2} \vs \textcircled{1}). This verifies that the sampling helps to learn a robust similarity metric for region matching. Adding the feature combination module (detailed in Sec. \ref{Sec4.1}) can bring a large improvement of 4.2 points in terms of EAO (\textcircled{3} \vs \textcircled{2}). This demonstrates the effectiveness of the proposed irregular dilated convolution module. It introduces a multi-scale modeling of target objects, without increasing much computation overhead. Furthermore, the object-aware classification (detailed in Sec. \ref{Sec3.2}) can also bring an improvement of 2.9 points in terms of EAO (\textcircled{4} \vs \textcircled{3}). This shows that the object-aware features generated by the proposed feature alignment module contribute significantly to the tracker. Finally, the tracker equipped with the plug-in online update module  (detailed in Sec. \ref{Sec4.2}) yields another improvement of 2.2 points (\textcircled{5} \vs \textcircled{4}), showing the scalability of the proposed framework.

\noindent\textbf{Feature combination.} We further evaluate the impact of dilated convolutions in the feature combination module and report the results on VOT-2018 in Tab. \ref{FCTAB}. The baseline setting is a normal convolution with dilation stride of $1$ along both the $X$-axis and $Y$-axis, \ie, $\Phi_{11}$. We observe that adding a standard convolution $\Phi_{11}$ brings an improvement of 0.8 points in terms of EAO (\textcircled{2} \vs \textcircled{1}). This indicates that the proposed parallel convolutions in the feature combination module is effective. It is very interesting to see that if we modify the dilation strides along $X$ and $Y$ directions to be different, the performance can be further improved, \eg, 1.3 points gains for \textcircled{3} \vs \textcircled{2} while 1.0 points gains for \textcircled{4} \vs \textcircled{2}. This verifies that the irregular dilations is effective to enhance feature representability. A combination of the three dilation kernels with different strides obtains the best results in our experiment. 
It is worth noting that the feature combination module with irregular dilations is extremely lightweight, and it has large potentials to be plugged into other models in both tracking and detection.


\begin{table}[!t]\selectfont
	\begin{minipage}[t]{.48\textwidth}
		
		\vspace{-6.2em}
		\setlength{\abovecaptionskip}{0pt}%
		\centering
		
		\fontsize{9.5pt}{4mm}
		\begin{threeparttable}
			
			\begin{tabular}{  @{}c@{}| @{}l@{} |@{}c@{}}
				
				\#Num & ~Settings & ~EAO~
				\\  \cline{1-3}
				\textcircled{1} & ~SiamRPN++~\cite{SiamRPN++}   & 0.414
				
				\\
				\textcircled{2} & ~Ocean tracker (ours)   & 0.455
				
				\\
				\textcircled{3} & ~$+$ GOT-10k in training ~ & 0.467
				
				\\
				\textcircled{4} & ~$+$ LaSOT in training ~  & 0.462
				
				

			\end{tabular}
		\end{threeparttable}
		\vspace{0.6em}
		\caption{Analysis of training settings on VOT-2018. It verifies that the main performance gains are induced by the model, rather than the additional data.}
		\label{SETTAB}

	\end{minipage}\hfill
	\begin{minipage}[t]{.48\textwidth}%

		\setlength{\abovecaptionskip}{0pt}%
		\centering
		\includegraphics[width=0.85\textwidth, height=0.37\textwidth]{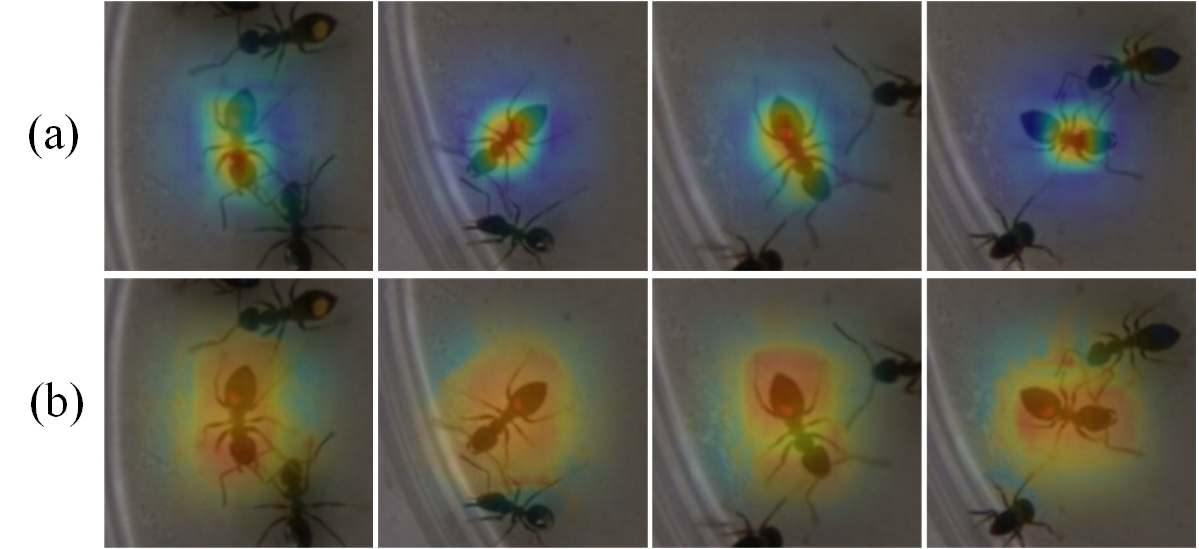} 
		\makeatletter\def\@captype{figure}\makeatother
		\caption{Visualization of (a) the regular-region feature and (b) the object-aware feature over the video ``ants1" in VOT-2018 \cite{VOT-2018} dataset. }
		\label{VISUALFIG}
		
	\end{minipage}
	\vspace{-3.2em}
	
\end{table}

\noindent\textbf{Training setting.} We conduct another ablation study to evaluate the impact of training settings. For a fair comparison, we follow the same setting as the well-performing SiamRPN++\cite{SiamRPN++}, \ie. training on YTB \cite{YTB}, VID \cite{ImageNet}, DET \cite{ImageNet} and COCO \cite{COCO} datasets for 20 epochs and using $6\times10^{5}$ image pairs in each epoch. As the results presented in Tab.~\ref{SETTAB} (\textcircled{2} \textit{v.s.} \textcircled{1}), our model surpasses SiamRPN++ by $4.1$ points in terms of EAO under the same training settings. Moreover, we further add GOT-10k~\cite{GOT10K} images into training, and observe that it brings an improvement of 1.2 points (\textcircled{3} \textit{v.s.} \textcircled{2}). This demonstrates that the main performance gains are induced by the proposed model. If we continue to add LaSOT\cite{LASOT} into training, the performance cannot improve further (\textcircled{4} \textit{v.s.} \textcircled{3}). One possible reason is that the object categories in LaSOT \cite{LASOT} have been covered by other datasets, thus it cannot further elevate model capacities.

\noindent\textbf{Feature visualization.} We visualize the features extracted before and after the alignment, \ie, the regular-region feature and object-aware feature, in Fig. \ref{VISUALFIG}.  We observe that the object-aware feature focuses on the entire object, while the regular-region feature concentrates on the center part of the target. The former improves the reliability of the classification since it provides a global view of the target. The latter contributes more to localize the object centerness since the features are more sensitive to local changes.

\section{Conclusion} \label{Sec6}
In this work, we propose a novel object-aware anchor-free tracking framework (Ocean) based upon the observation that the anchor-based method is difficult to refine the anchors whose overlap with the target objects are small. Our model directly regresses the positions of target objects in a video frame instead of predicting offsets for predefined anchors.  Moreover, the learned object-aware feature by the alignment module provides a global description of the target, contributing to the reliable matching of objects. The experiments demonstrate that the proposed tracker achieves state-of-the-art performance on five benchmark datasets. In the future work, we will study the update of the parameters in the object-aware classification network without integrating an additional online branch. Besides, applying our framework to other online video tasks, \eg, video object detection and segmentation, is also a potential direction.


\clearpage

%
%
\bibliographystyle{splncs04}
\bibliography{egbib}

\clearpage

 --------------------- {\large \textbf{Supplementary Material}} -----------------------
\\

\noindent The supplementary material presents the additional experiments of Section $5$:

\noindent 1) We provide additional ablation experiments. 

\noindent 2) We provide qualitative comparisons of our tracker with state-of-the-arts.

\subsubsection{Impact of the convolution layers in the anchor-free networks.} For both of the regression network and regular-region classification network 
(\emph{i.e.}, ``Conv'' of the anchor-free networks in the Fig. \ref{FRAMEWORKFIG}, we use four convolution layers to predict the fine-grained regression and classification score maps.
To study the impact of the number of convolution layers in these two networks,  we perform an ablation experiment on OTB-100 \cite{OTB-2015}, as shown in Tab. \ref{LAYTAB}. We can see that as the number of convolution layers increases, the performance (\emph{i.e.}, AUC) becomes saturated. The balanced choice between performance and speed is 3 or 4 in our model. 

\begin{table}
	\centering
	\vspace{-1em}
	\fontsize{9pt}{4mm}\selectfont
	\begin{threeparttable}
		\begin{tabular}{ @{}c@{} | @{}c@{} @{}c@{} @{}c@{} @{}c@{} @{}c@{} @{}c@{}}
			
			The number of layers~& ~~~0~~~ & ~~~1~~~ & ~~~2~~~ & ~~~3~~~ & ~~~4~~~ & ~~~5~~~
			\\  \cline{1-7}
			~~Performance (AUC $\uparrow$)~~ &~~ 0.645 ~~&~~ 0.657~~ & ~~0.665~~ & ~~0.672~~  & ~~0.673~~ & ~~ 0.670 ~~
		\end{tabular}
	\end{threeparttable}
	\vspace{0.5em}
	\caption{Impact of the number of layers in the anchor-free networks.}
	\vspace{-2.5em}
	\label{LAYTAB}
\end{table}

\noindent\textbf{Analysis of rectification capacity.} The rectification capacity indicates the prediction accuracy ($mIoU$) of the model when the tracker drifts from the target.  To evaluate the rectification capacity, we first sample exemplar and search image pairs from adjacent frames in the VOT-2018 dataset \cite{VOT-2018}.  We sample locations on the score maps away from the target to simulate weak predictions, and the shifting magnitude is illustrated in the first row of Tab. \ref{RETTAB}. Then we compute the overlap between the predicted bounding box and the groundtruth,  \emph{i.e.}, \emph{mIoU} in Tab. \ref{RETTAB}. Larger \emph{mIoU} means that the regression network can better rectify the inaccurate prediction. We can see that the performance (\emph{mIoU}) of the proposed model outperforms SiamRPN++ \cite{SiamRPN++} when the tracker drifts away from the target's center.  This demonstrates the superior robustness of our tracker compared to the anchor-based method.

\begin{table}\footnotesize
	\begin{center}
		
		\vspace{-1em}
		\fontsize{8pt}{4mm}
		\begin{threeparttable}
			\begin{tabular}{ @{}c@{} | @{}c@{} @{}c@{} @{}c@{}  @{}c@{} @{}c@{} @{}c@{} @{}c@{} @{}c@{}}
				\emph{Distance} (\emph{pixels}) ~  & ~~~~8~~~~ & ~~~~16~~~~ & ~~~~24~~~~ & ~~~~32~~~~ & ~~~~40~~~~ & ~~~~48~~~~ & ~~~~56~~~~
				\\
				\cline{1-9} 
				~~ \emph {mIoU} of SiamRPN++ \cite{SiamRPN++}~~& 0.65 & 0.48 & 0.36 & 0.30 & 0.28 & 0.25 & 0.21 & 
				\\
				~~ \emph{mIoU} of our tracker ~~& 0.73 & 0.73 & 0.72 & 0.71 & 0.71 & 0.72 & 0.54

			\end{tabular}
			
		\end{threeparttable}
		\vspace{0.3em}
		\caption{Comparisons of rectification capacity between SiamRPN++ and our model. \emph{mIoU} indicates mean \emph{IoU}. \emph{Distance} indicates shifting magnitude ($\ell_1$ distance) for generating search images.}

		\vspace{-3em}
		\label{RETTAB}
	\end{center}
\end{table}

\begin{figure}[!t]
	\begin{center}
		\vspace{-0.5em}
		
		\includegraphics[height=1\linewidth]{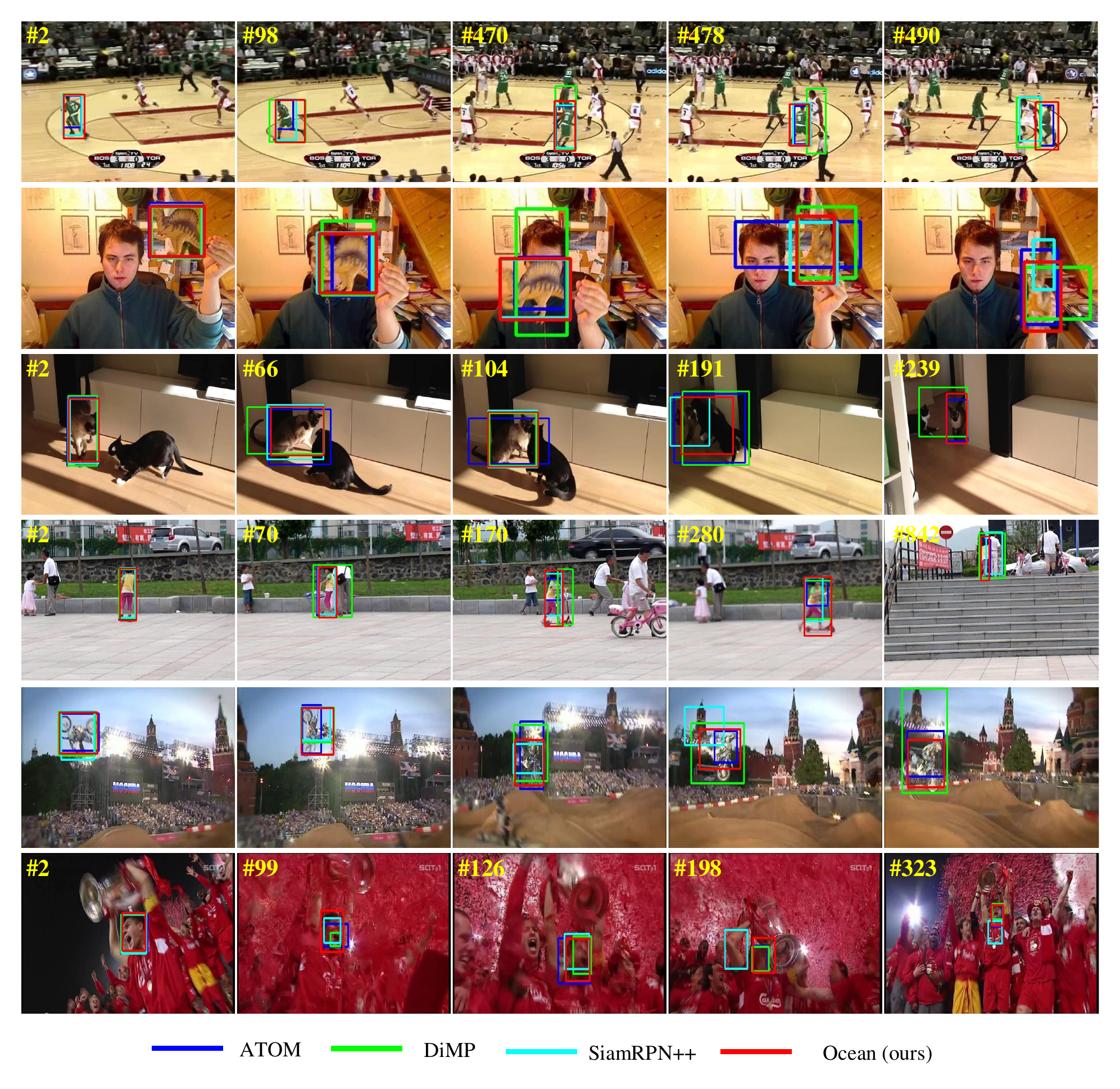}
		\vspace{-2.5em}
		\caption{Visual comparisons of our tracker with statr-of-the-art trackers on 6 video sequences:
			\emph{basketball, dinasour, fernando, girl, motocross1 and soccer.}}

		\label{FIG1}
	\end{center}
	\vspace{-2.7em}
\end{figure}

\noindent\textbf{Qualitative comparisons.} 
Fig. \ref{FIG1} qualitatively compares the results of recent top-performing trackers: DiMP \cite{DiMP}, ATOM \cite{ATOM}, SiamRPN++ \cite{SiamRPN++} and our Ocean tracker on 6 challenging sequences. SiamRPN++ drifts from the target when fast motion occurs (\emph{soccer}). The underlying reason is that fast motion results in imprecise prediction, which is difficult to be rectified by the anchor-based method. By contrast, 
our model performs better in this case, since it is capable of rectifying inaccurate bounding boxes and robust to noisy predictions. 
This verifies the advancement of our anchor-free regression mechanism compared to anchor-based methods. 
When the target undergoes large deformation or rotation, \emph{e.g.}, \emph{dinasour} and \emph{motocross1},  the predicted locations of ATOM and DiMP are not accurate enough. One of the reasons is that the regular-region sampling strategy in these approaches may lack global information to generate discriminative appearance features. Benefiting from the object-aware feature, our model can predict better results in this case. The visualization results demonstrate the robustness and effectiveness of the proposed tracker. 

\end{document}